\documentclass{article}

\usepackage[preprint]{neurips_2023}

\usepackage[utf8]{inputenc} 
\usepackage[T1]{fontenc}    
\usepackage{hyperref}       
\usepackage{url}            
\usepackage{booktabs}       
\usepackage{amsmath}        
\usepackage{amsfonts}       
\usepackage{nicefrac}       
\usepackage{microtype}      
\usepackage{xcolor}         
\usepackage{graphicx}
\title{Detecting Concept Drift in Neural Networks Using Chi-squared Goodness of Fit Testing}

\begin{document}

\maketitle

\begin{abstract}
As the adoption of deep learning models has grown beyond human capacity for verification, meta-algorithms are needed to ensure reliable model inference. Concept drift detection is a field dedicated to identifying statistical shifts that is underutilized in monitoring neural networks that may encounter inference data with distributional characteristics diverging from their training data. Given the wide variety of model architectures, applications, and datasets, it is important that concept drift detection algorithms are adaptable to different inference scenarios. In this paper, we introduce an application of the $\chi^2$ Goodness of Fit Hypothesis Test as a drift detection meta-algorithm applied to a multilayer perceptron, a convolutional neural network, and a transformer trained for machine vision as they are exposed to simulated drift during inference. To that end, we demonstrate how unexpected drops in accuracy due to concept drift can be detected without directly examining the inference outputs. Our approach enhances safety by ensuring models are continually evaluated for reliability across varying conditions.
\end{abstract}

\section{Introduction}

In the rapidly evolving landscape of deep learning (DL), the challenge of concept drift detection (CDD) has gained prominence as neural network models encounter inference data that deviates from their training data \cite{baier2022detecting}. CDD algorithms are specifically designed to identify potential shifts in the distribution parameters of random processes over time, such cases are referred to as non-stationary random processes\cite{iwashita2018overview, 8496795, 7296710}. Given the diversity of DL model architectures \cite{fukushima1980neocognitron, vaswani2017attention, rosenblatt1958perceptron, 6795724} and domain applications, the preference is for CDD algorithms that exhibit flexibility across various configurations. Furthermore, meta-algorithms monitoring these neural networks during inference ideally require minimal human-in-the-loop interaction due to time and budget constraints. To meet these preferences, our approach leverages the $\chi^2$ Goodness of Fit (GoF) Hypothesis Testing on neural network hidden layer outputs, known as activations, to detect artificially induced drift on the MNIST dataset \cite{deng2012mnist}.

In data-intensive fields like machine vision, a common challenge with hypothesis testing is the risk of accepting a statistically significant alternative hypothesis with large sample sizes yielding small p-values \cite{lin2013}. To address this challenge, we conduct numerous smaller tests generated from randomized subsets of the activations, which are subsequently aggregated. The decision to use the $\chi^2$ Goodness of Fit (GoF) test is motivated by its suitability for detecting statistical shifts. This choice finds support in the work of the National Institute of Standards and Technology (NIST), where the test is deployed to detect statistical shifts in pseudorandom number generators \cite{turan2018recommendation}. Similar work using $\chi^2$ GoF testing for CDD on time-series data can be found in the Alibi-detect library \cite{Van_Looveren_Alibi_Detect_Algorithms_2023}.

\section{Background}
\subsection{Concept Drift Detection}
\begin{figure}
  \centering
  \includegraphics[scale=0.50]{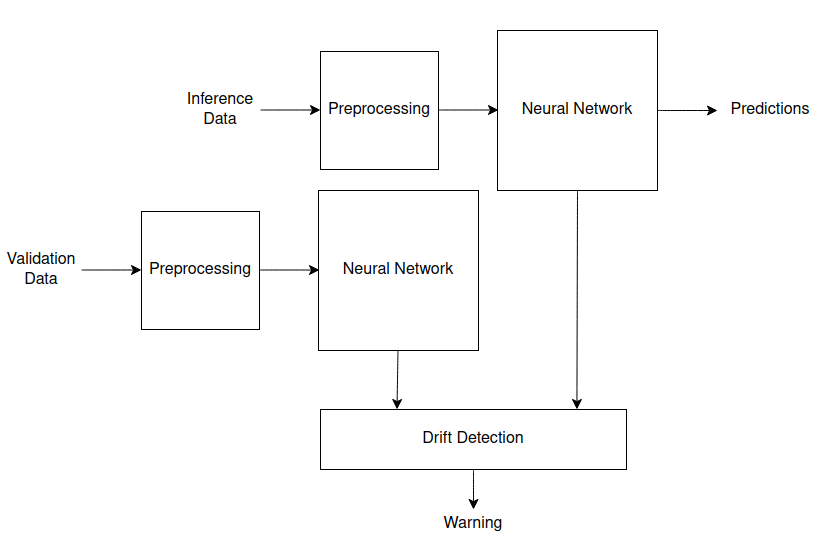}
  \caption{Concept Drift Detection for Deep Learning System Diagram}
  \label{drift-diagram}
\end{figure}

CDD falls under the umbrella of Concept Drift that is defined as a shift in the underlying distribution of data during inference as time progresses. Those familiar with random processes will recognize this as recognizing non-stationary random processes where the underlying distribution across samples taken from interval ($t_{0}$, $t_{1}$) is not the same as if they were taken from ($t_{0}$ + a, $t_{1}$ + a) \cite{leongarcia08}. In the context of DL, the shift in the underlying distribution commonly occurs as we train a network on a dataset, evaluate it on data that is similar to the training data, and then deploy said network where it encounters inference data that varies from the training and evaluation datasets. A system diagram of this setup can be seen in Figure \ref{drift-diagram} 
\subsection{Hypothesis Testing}
Hypothesis testing is a formal statistical procedure that involves comparing new data, obtained from a sample space denoted as $\Omega$, with existing data and/or parameters derived from $\Omega$. The process aims to assess whether observed differences or patterns are statistically significant, when compared to a-prior expectations of $\Omega$. 

Assuming data has been collected, the general process of two-tailed Hypothesis Testing is as follows - 
\begin{enumerate}
    \item Formulate the Null Hypothesis $H_{0}$ that defines prior assumptions of $\Omega$ and the Alternative Hypothesis $H_{A}$ that defines a potential deviation from the status-quo. 
    \item Establish a significance level $\alpha$ that serves as the probability of a Type I error of rejecting $H_{0}$ when it is in fact correct.
    \item Compute the test statistic \textit{T} for $H_{A}$ we call \textit{$T_{A}$} from experimental observations. Typically this is derived from a statistical parameter that defines the underlying probability density function (PDF) behind the experiment.
    The expected values can be from prior experiments or derived directly from a PDF believed to define $\Omega$
    \item Compare the test statistic $T_{A}$ to the critical region test statistic for $H_{0}$ we call $T_{O}$ defined by the sample size and the significance level and typically found through a lookup table. We then determine whether to accept ($|T_{O}| \geq |T_{A}|$) or reject ($|T_{O}| < |T_{A}|$) $H_{0}$.
\end{enumerate}
\subsubsection{Chi-squared Goodness of Fit Test}

The Chi-squared ($\chi^2$) test statistic was initially developed by Karl Pearson to quantitatively compare experimental observations with theoretical expectations. Pearson's early applications included analyses such as demonstrating the outcomes of roulette tables and coin flips, comparing them against Gaussian and binomial distributions respectively \cite{FRS2009XOT}. Ronald Aylmer Fisher expanded on Pearson’s work by defining the degrees of freedom for the test ($df = category\_count - 1$) and the probability density function of the $\chi^2$ distribution \cite{fisher1923statistical}. This framework was notably applied to confirm that radioactive decay follows a Poisson distribution, showcasing the test’s utility in empirical research \cite{Rutherford_Chadwick_Ellis_2010}.

The general form of the $\chi^2$ GoF Test follows from the aforementioned general Hypothesis Testing outline of an expected sample space $\Omega$ where the test statistic defined as - 

\[\textit{T} = \chi^{2} = \Sigma\frac{(observed-expected)^{2}}{expected}\]
\subsubsection{Vanishing P-Value Problem}
The vanishing p-value problem is an effect of increased sensitivity of hypothesis tests to large sample sizes, leading to very small deviations from the null hypothesis leading to "statistically significant" p-values \cite{lin2013}. This phenomena can be attributed to the weak law of large numbers that states that as more sample are collected from a sample space $\Omega$, the sample mean will converge to the population mean that defines $\Omega$. Consequently, with large sample sizes, the statistical power of a test increases, making it more likely to detect even trivial differences as significant, rendering hypothesis testing worthless for many big data applications.

\section{Methodology}
All of the experiments were performed using a mixture of the following Python packages: Pytorch, Numpy, Scipy, and Pandas. The hardware used has an AMD Ryzen Threadripper 3970x processor and an Nvidia 4070 graphics card.
\subsection{Data}
All models underwent training using the widely adopted MNIST dataset \cite{deng2012mnist}, a standard benchmark in machine vision renowned for its collection of handwritten digit images. The dataset comprises a total of 70,000 images, with 60,000 allocated for training and 10,000 for testing. To facilitate our drift experiments, we strategically employed a 5-1 training-validation split. This division ensured a dedicated validation subset, equal in size to the test data, which serves as an expected dataset representative of the distribution observed during training.
\subsection{Drift}
\begin{figure}
  \centering
  \includegraphics[scale=0.4]{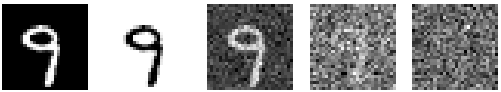}
  \caption{Drift Demonstration - left to right: baseline, inverted, AGN with $\sigma$ 0.3, 1.2, and 4.5}
  \label{drift}
\end{figure}

Two forms of simulated drift were applied to all of the test data prior to inference by the neural networks. The first form of drift consisted of inverting the images by subtracting each image from the max value of each batch. The second form of drift involves Added Gaussian Noise (AGN) with a mean of zero and standard deviations [0,4.5] in increments of 0.15. Both forms of drift are demonstrated in Figure \ref{drift}.  
\subsection{Training}
All models were trained using the same configuration with the exception of the number of epochs, which was 18 for the MLP, 8 for the CNN, and 36 for the ViT. These epochs were chosen based on achieving a validation accuracy of approximately 90\%.The aforementioned MNIST training set of 50,000 images were split into batches of 32. Cross-entropy was used for the loss function and stochastic gradient descent was used as the criterion for backpropagation with a learning rate of $10^{-3}$.
\subsection{Models and Activations}
\subsubsection{Multi-layer Perceptron}
The MLP model used consisted of 4 fully connected hidden layers with 256 nodes. All of the hidden layers were followed by a ReLu activation function. All of the hidden layer output activations were used for the $\chi^{2}$ GoF tests. 
\subsubsection{Convolutional Neural Network}
The CNN model used consisted of 2 convolutional hidden layers and 2 fully connected hidden layers. The first convolutional layer consisted of 16 3x3 kernels that was then subject to a 2x2 max-pooling operation, followed by a ReLu activation function. The second convolutional layer consisted of 32 3x3 kernels followed by a ReLu activation. The first fully connected layer has an output size of 32 while the second fully connected layer has an output size of 84. Both were exposed to ReLu activations. All of the convolutional and fully connected layer outputs activations were used for the $\chi^{2}$ GoF tests. The CNN used was derived from a PyTorch tutorial \cite{pytorch-tutorial}.
\subsubsection{Vision Transformer}
The CNN model used consisted of 6 encoding blocks with 8 attention heads, and patch sizes of 7. The fully-connected head of each encoder block had a hidden layer size of 128. All of the keys, queries, and values, attention heads, and the two feed-forward blocks' output activations from each of the encoder blocks were used for the the $\chi^{2}$ GoF tests. The ViT model used was derived from this kaggle competition submission \cite{kaggle-vit} which was based on work done by Google Brain \cite{dosovitskiy2020image}.
\subsection{Chi-squared Goodness of Fit Tests}
\begin{figure}
  \centering
  \includegraphics[scale=0.4]{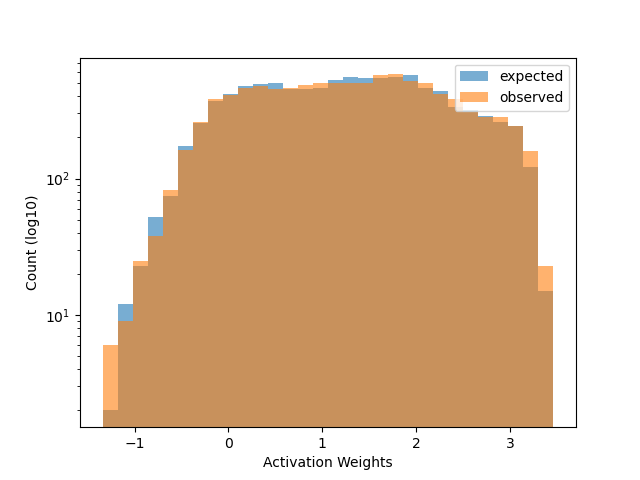}
  \includegraphics[scale=0.4]{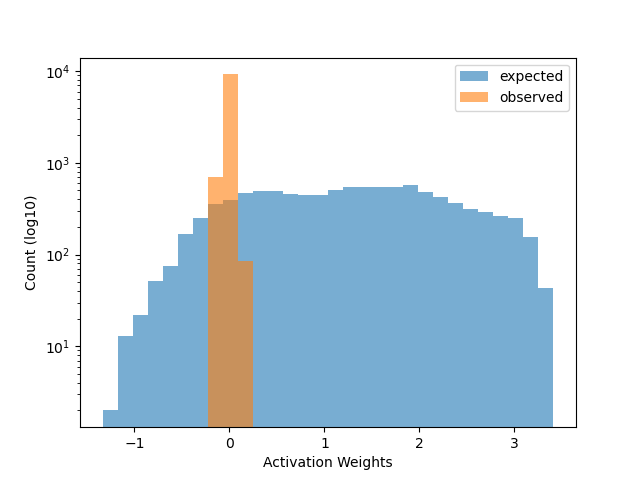}
  \caption{Histogram Activations MLP Example - expected validation data compared to observed test data (left) and expected validation data compared to observed inverted test data (right)}
  \label{obs-exp-hist}
\end{figure}

In order to compare the expected activations, derived from the validation dataset across all networks, to the observed activations, whether they be the baseline test set, subject to pixel inversion, or AGN, all of the individual activation notes were written to files. For each node, we create randomized subsets of size 500 for both the expected and observed values. In the case where we have a test set of 10,000 datapoints, that totals to 20 hypothesis tests per node. We find the minimum and maximum values between expected and observed subsets in order to find the histogram bounds. From there, 30 bins are created uniformly between said bounds. A visualization between observed and expected histograms is shown here in Figure \ref{obs-exp-hist}. The $\chi_A^{2}$ test statistic is then computed by comparing each of the expected and observed histogram bin frequencies. From there, we compare the test statistic to the critical region determined by a confidence level of 0.99 and 29 degrees of freedom ($histogram\_bin\_count - 1$) yielding $\chi_A^{2}$ = 49.588 to decide whether the current observed subset fits the same distribution as that of the expected distribution. Once all of the hypothesis tests have been completed across all the subsets for a node, we compute an aggregated pass rate of how many times the null hypothesis was upheld divided by the total number of subset tests. Once all of the pass rates have been computed across all of the activations across a network, we compute a global pass rate by setting a threshold of 0.5 and counting the number of nodes had an aggregated pass rate above said threshold and dividing by the total number of nodes.

\section{Results}
\begin{figure}
  \centering
  \includegraphics[scale=0.40]{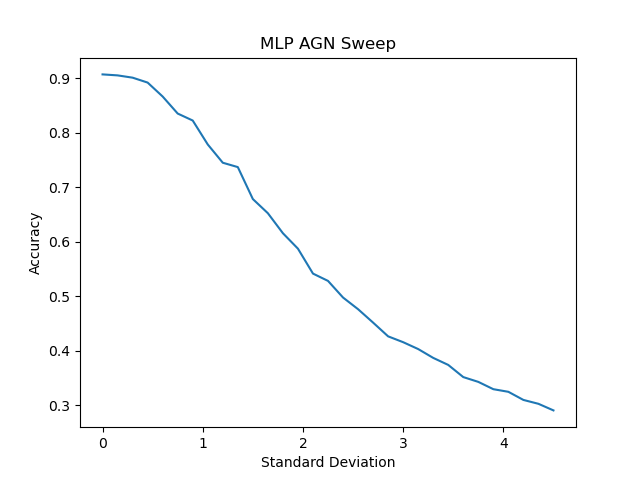}
  \includegraphics[scale=0.40]{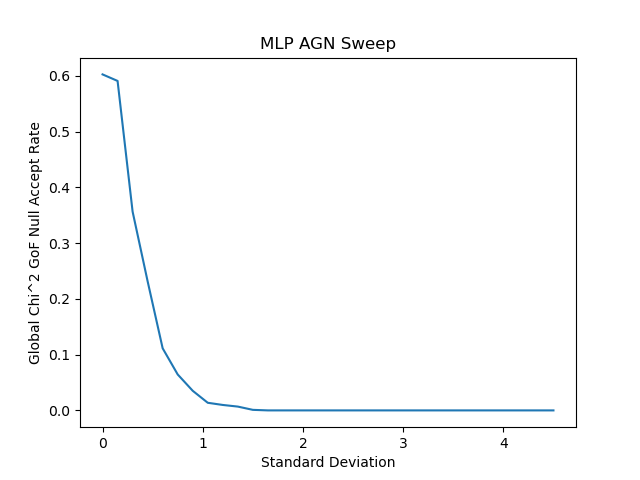}
  \includegraphics[scale=0.40]{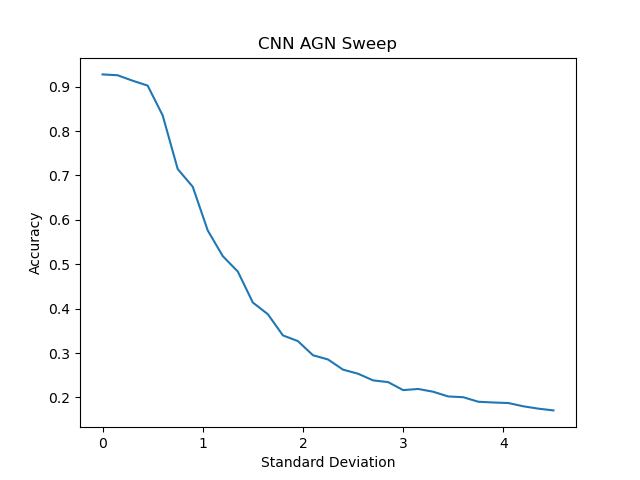}
  \includegraphics[scale=0.40]{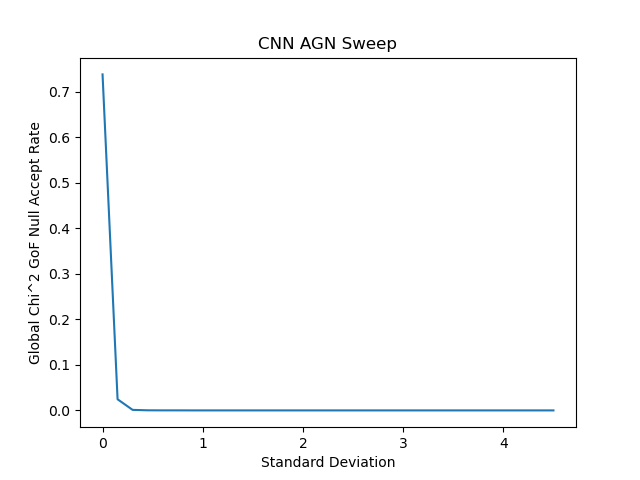}
  \includegraphics[scale=0.40]{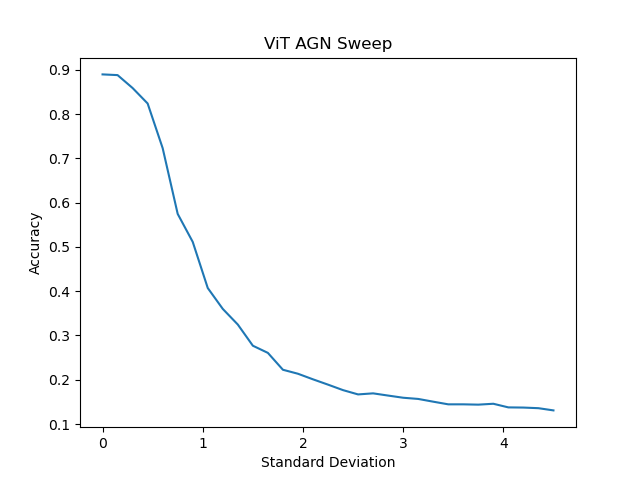}
  \includegraphics[scale=0.40]{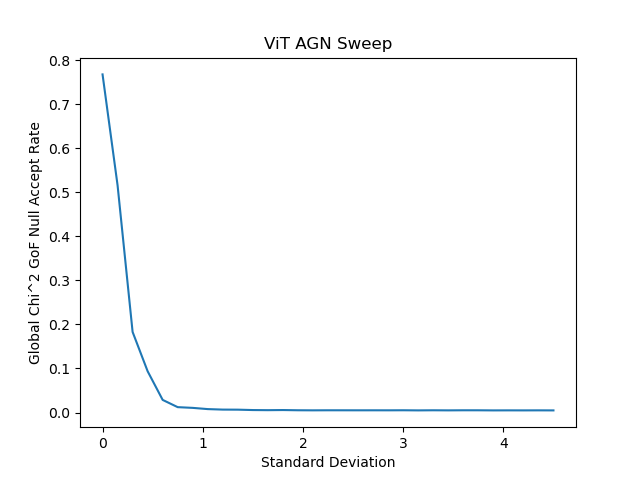}
  \caption{Accuracy vs. Standard Deviation for all Networks (left) and $\chi^{2}$ GoF Test Pass Rate vs. Standard Deviation for all Networks (right)}
  \label{AGN-accuracy}
\end{figure}

The accuracies of all of the models across the validation, test, and test after inversion can be found in Table  \ref{accuracy-table}. Furthermore, we monitor accuracy with respect to incremental increases in AGN across all networks in Figure  \ref{AGN-accuracy}. The global pass rates for the inversion experiments can be seen in Table  \ref{accuracy-table}. The global pass rates for all of the AGN sweeps across all networks can be seen in Figure \ref{AGN-accuracy}.
\begin{table*}
  \caption{Neural Network Baseline and Inverted Accuracies and Inverted Hypothesis Testing Results}
  \label{accuracy-table}
  \centering
  \begin{tabular}{lcccccc}
    \toprule
    Architecture & Val & Test & Inv. Test & Test $\chi^{2}$ GoF & Inv. Test $\chi^{2}$ GoF \\
    & Acc. & Acc. & Acc. & Pass Rates & Pass Rates \\
    \midrule
    MLP & 0.905 & 0.907 & 0.054 & 0.603 & 0.002    \\
    CNN & 0.919 & 0.928 & 0.107 & 0.738 & 0.005    \\
    ViT & 0.890 & 0.890 & 0.097 & 0.768 & 0.004\\
    \bottomrule
  \end{tabular}
\end{table*}
\section{Conclusion}
The experiments in applying the $\chi^{2}$ GoF test to neural network hidden layer activations demonstrate that the hypothesis tests pass at lower rates when various kinds of neural networks are exposed to concept drift at inference time. In particular, our results in Figure \ref{AGN-accuracy} reveal a strong correlation between model accuracy and GoF pass rates across all network architectures tested, with both metrics declining in tandem as the drift intensity increases. This clear relationship confirms the effectiveness of our approach as a reliable indicator of model performance degradation. The consequence of this correlation is that the $\chi^{2}$ GoF test can be used as a simple heuristic to warn relevant parties when a neural network is exposed to data that it has not been prepared to handle without actively looking at the inference outputs. In other words, those in charge of deploying a neural network can use hypothesis testing to notify them in the event that the model has unintended drops in accuracy with minimal effort. Our hope is that this approach can be used to assist groups in managing neural networks at scale that have consequences on people's daily lives.
\section{Future Work}
Given the preliminary promising results of detecting neural network inference drift, the natural next steps would be to work our way towards concept drift understanding where we attempt to quantify not only that something is going wrong but to identify what is causing the drift. With current experiments, inversion simulates "sudden drift", while AGN sweep simulates "incremental drift" \cite{8496795}. It seems within the realm of possibility that another algorithm could process the $\chi^2$ GoF failure rates at the node or layer level to identify these different forms of drift. The ultimate goal would be to see if a system could be developed to adapt to drift with minimal human interaction.

A significant practical challenge in our current approach is the substantial storage requirements for saving a large number of activation values across all network layers. For production systems with millions of parameters, this storage overhead could become prohibitive. Future work should explore more storage-efficient alternatives, such as maintaining only statistical summaries (e.g., sufficient statistics) of the expected activation distributions rather than storing raw activation values. This would dramatically reduce memory requirements while still enabling drift detection.

There seems to be an opportunity to treat the parameters of the hypothesis itself, such as the sample size and the significance level to try and maximize correlation between the GoF pass rate and the accuracy of a given model. We do not do that in this paper, as the eventual methodology and final parameters would be unlikely to generalize across applications. In our experimental setup, we sampled the actual activation floating point values from an expected validation dataset and compared those to observed samples to perform the hypothesis test. An alternative approach would be to derive statistical distribution parameter estimates such as sample mean and sample standard deviation from the expected distribution and instead compute the expected estimates of the number of samples in each bin to those observed. This would certainly be less computationally expensive for more resource-scarce environments. All of the networks used in this paper were trained for machine vision on MNIST, and it would be interesting to see if the drift detection results translated into different tasks and data, particularly with transformers trained for Natural Language Processing.
\section{Additional Materials}
Due to corporate data policies and the transition of research environments, the codebase for this work is not publicly available in a repository. However, researchers interested in implementation details or specific aspects of the methodology are encouraged to contact the corresponding author for further discussion and potential collaboration.
\begin{figure}
  \centering
  \includegraphics[scale=0.5]{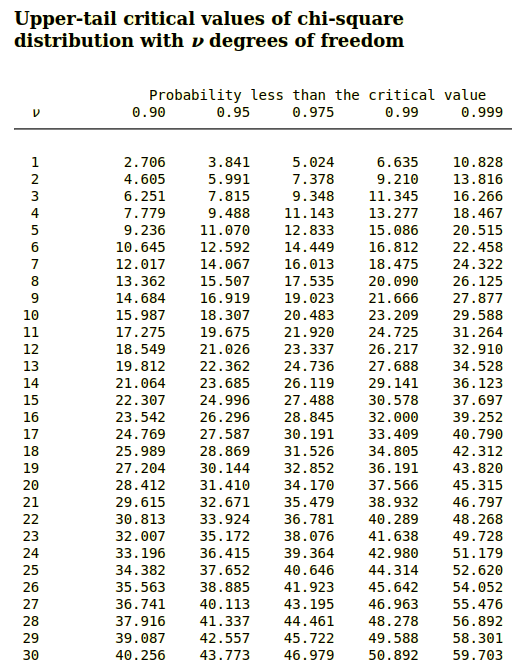}

  \caption{NIST $\chi_A^2$ table (probability less than the critical value ==> 1 - $\alpha$)}
  \label{NIST-chi-table}
\end{figure}
\section{Acknowledgements}
We thank Dr. Jin Cao for her expert feedback on the initial proof of concept of this work.

\bibliographystyle{plainnat}\bibliography{bibliography}

\end{document}